\documentclass[numeric]{article}

\usepackage{lineno,hyperref}

\usepackage[square]{natbib}

\usepackage{amsmath}
\usepackage{amsfonts}
\usepackage{graphicx,graphics,color,epsfig}
\usepackage{amsmath}
\usepackage{amssymb,amsthm}
\usepackage{algorithm}
\usepackage{algorithmic}  
\usepackage[algo2e]{algorithm2e}
\usepackage{pgf, tikz, pgfplots}
\usetikzlibrary{intersections}
\usepgfplotslibrary{fillbetween}
\usepackage{subcaption}

\usepackage{geometry}
\definecolor{darkblue}{rgb}{0,0,0.5}

\hypersetup{colorlinks=true,linkcolor=blue,citecolor=blue,urlcolor=blue} 

\newcommand{\newref}[2]{\hyperref[#2]{#1~\ref*{#2}}} 

\usepackage{multicol}\columnseprule 0.4pt\raggedcolumns

\newtheorem{theorem}{Theorem}[section]

\newtheorem{proposition}[theorem]{Proposition}

\graphicspath{{./figures/}}

\begin{document}

\title{R2L: Reliable Reinforcement Learning:  \\ Guaranteed Return \& Reliable Policies in Reinforcement Learning}

\author{Nadir Farhi \\ Cosys-Grettia, Univ Gustave Eiffel, F-77454 Marne-la-Vallée, France. \\ \textcolor{blue}{\texttt{nadir.farhi@univ-eiffel.fr}}}

\date{}

\maketitle

\begin{abstract}
In this work, we address the problem of determining reliable policies in reinforcement learning (RL), with a focus on optimization under uncertainty and the need for performance guarantees. While classical RL algorithms aim at maximizing the expected return, many real-world applications—such as routing, resource allocation, or sequential decision-making under risk—require strategies that ensure not only high average performance but also a guaranteed probability of success. To this end, we propose a novel formulation in which the objective is to maximize the probability that the cumulative return exceeds a prescribed threshold. We demonstrate that this reliable RL problem can be reformulated, via a state-augmented representation, into a standard RL problem, thereby allowing the use of existing RL and deep RL algorithms without the need for entirely new algorithmic frameworks. Theoretical results establish the equivalence of the two formulations and show that reliable strategies can be derived by appropriately adapting well-known methods such as Q-learning or Dueling Double DQN. To illustrate the practical relevance of the approach, we consider the problem of reliable routing, where the goal is not to minimize the expected travel time but rather to maximize the probability of reaching the destination within a given time budget. Numerical experiments confirm that the proposed formulation leads to policies that effectively balance efficiency and reliability, highlighting the potential of reliable RL for applications in stochastic and safety-critical environments.
\end{abstract}

\textbf{Keywords: }
Reinforcement learning, reliable policies, optimal control, reliable routing.

\section{Introduction}

Reinforcement Learning (RL) offers a powerful framework for training autonomous agents to make sequential decisions in complex and uncertain environments. Formally grounded in the theory of Markov Decision Processes (MDPs), RL seeks to learn policies that maximize expected cumulative rewards through interaction with stochastic dynamics. Unlike traditional control methods that rely on explicit system models, RL enables data-driven learning, both in model-based and model-free settings. Despite its success in domains such as robotics, autonomous driving, and games, RL remains challenged by the exploration–exploitation dilemma, sample inefficiency, approximation errors, and the risk of converging to sub-optimal or unsafe policies. These limitations have motivated a growing body of research on more robust, data-efficient, and theoretically grounded approaches.

While classical RL focuses on optimizing the expected return, many real-world applications demand stronger guarantees on performance under uncertainty. In safety-critical settings, maximizing expectation alone is insufficient; instead, one often seeks to ensure that performance remains above a prescribed level with high probability, or to optimize risk-sensitive criteria such as Conditional Value-at-Risk (CVaR) or the probability of success. For example, in finite-horizon problems, one may aim to find a policy $\pi$ that maximizes the probability $P_{\pi}(G_T \geq \rho)$, where $G_T$ is the cumulative return over time horizon $T$ and $\rho$ is a performance threshold. This notion of reliable reinforcement learning (R2L) thus extends beyond expectation-optimal policies, incorporating guarantees of stability, robustness, and safety that are indispensable in domains such as autonomous driving, healthcare, industrial automation, and cyber-physical systems.

Reliable RL can be viewed as part of a broader family of approaches that include safe RL and robust RL, which explicitly account for uncertainty, variability, and risks in sequential decision making. These frameworks draw on concepts from risk-sensitive control, robust optimization, and distributional RL, and they aim to produce policies that are not only effective on average but also resilient to stochastic disturbances, distribution shifts, or adversarial perturbations. By emphasizing probabilistic guarantees and minimizing the likelihood of catastrophic failures, reliable RL bridges reinforcement learning, robust control, and stochastic optimization, making it particularly relevant for safety-critical, real-world applications.

In this article, we propose a new formulation of the reliable RL problem, where the objective is to maximize the probability that the cumulative return exceeds a given threshold. Our contribution is a state-augmented formulation, in which the state space is extended to include not only the original system variables but also the remaining return required to reach the threshold. This embedding enables the reliable RL problem to be recast as a standard RL problem, solvable using adapted versions of existing algorithms. In this formulation, the temporal difference (TD) and Bellman equations are rewritten in terms of probabilities of threshold satisfaction rather than expected returns, providing a principled and tractable framework for deriving reliable policies. To illustrate the approach, we apply it to the problem of reliable routing, where the goal is not to minimize expected travel time but to maximize the probability of arriving at a destination within a specified time budget. In the stochastic shortest-path literature, this is known as the Stochastic On-Time Arrival (SOTA) problem.

The main contributions of this work are as follows:
\begin{itemize}
  \item We propose a state-augmented approach that reformulates the problem of maximizing the probability of exceeding a minimum total   return as a new RL problem, which can be solved with standard RL techniques.
  \item We situate our approach in relation to existing methods—particularly distributional RL—clarifying both the similarities and the key differences.
  \item We demonstrate the approach on the problem of reliable routing, showing how maximizing the probability of on-time arrival can be formulated and solved within our framework.
\end{itemize}

The remainder of this article is organized as follows. Section 2 reviews related work, including risk-sensitive RL, exponential utility models, cumulative prospect theory, coherent risk measures, min–max criteria, quantile-based criteria, dynamic risk measures, and distributional RL. Section 3 provides preliminaries, reviewing RL concepts and notations and summarizing risk-sensitive control approaches. Section 4 introduces our proposed approach, Reliable Reinforcement Learning (R2L), presenting the problem formulation, state-augmented representation, and associated algorithms. Section 5 illustrates the approach through the reliable routing problem, with results and interpretations. Section 6 concludes with a discussion of limitations and future research directions. 

\section{Related works}

The development of reinforcement learning (RL) has its origins in the seminal Q-learning algorithm \citep{W89}, which introduced the state–action value function as a cornerstone of sequential decision making. Its theoretical convergence was subsequently established in a series of works~\citep{W92, Jaa94, Tsi94, BT96}, providing the foundation for subsequent advances. With the integration of neural networks, RL entered the era of deep reinforcement learning, where algorithms such as the Deep Q-Network (DQN) and its extensions—including Double DQN \citep{Has16}, Dueling DQN \citep{Wang16}, prioritized replay variants \citep{Sch15}, and the comprehensive Rainbow architecture \citep{Hes18}—demonstrated remarkable empirical success. In parallel, policy-gradient methods matured from the early REINFORCE algorithm \citep{Wil92} to more sophisticated trust-region and clipping-based approaches such as TRPO \citep{SSJP15} and PPO \citep{SW17}, which today serve as the backbone of many applied RL systems.

The question of how to incorporate risk and uncertainty into sequential decision making has long been studied, beginning with Markowitz’s mean–variance framework for portfolio selection \citep{Mar52} and its extensions \citep{Ste01}. Exponential utility functions \citep{Arr71, HM72} and behavioral models such as cumulative prospect theory (CPT) \citep{TK92} further enriched the theoretical landscape. Within the context of Markov decision processes (MDPs), the computational complexity of variance-based optimization was established in \citep{Fil89, MT13}, while coherent risk measures \citep{Art99} and chance constraints \citep{DR03} provided principled formulations of risk aversion. More recent surveys \citep{PF21, Nel21} have synthesized these developments under the umbrella of risk-sensitive reinforcement learning, highlighting both theoretical contributions and practical applications.

Building on these foundations, several algorithmic extensions of RL have been proposed to explicitly address risk-averse and robust control. Early contributions include min–max Q-learning for worst-case guarantees \citep{Heg94} and Q-learning extensions to recursive risk measures \citep{MN02, She13}. Quantile-based methods \citep{Fil95, Gil17, Li22} and simulation-driven approximate dynamic programming approaches \citep{JP18} further advanced the modeling of risk in sequential settings, while the notion of dynamic risk measures for MDPs was formalized in \citep{Rus10} and extended in \citep{Cho17}.

In parallel, the fields of robust and safe reinforcement learning have gained prominence as practical approaches to reliability. Robust dynamic programming, introduced by \citep{Iyen05}, provided the mathematical foundation for robustness in sequential decisions, and subsequent work \citep{MD05} established early robust RL formulations. Comprehensive surveys such as \citep{JKH22} have since organized robust RL methods according to the sources of uncertainty (transitions, disturbances, actions, or observations). Complementary research on safe RL has been summarized in the influential survey by \citep{GF15}, while more recent contributions \citep{GLY24} examine safe RL across multiple dimensions, including theory, empirical evaluation, and application. A broad synthesis of both safe and robust RL is provided in \citep{TR24}, which also emphasizes ethical considerations and proposes practical guidelines for deployment.

A related line of research, known as distributional reinforcement learning, has introduced a paradigm shift by modeling the full distribution of returns rather than only their expectation. The pioneering work of \citep{Bell17} established distributional value iteration, later extended in \citep{Ly19}, and demonstrated significant empirical improvements. Beyond their algorithmic appeal, distributional methods are particularly well suited to reliable decision making, as they provide access to risk-sensitive criteria such as Value-at-Risk and Conditional Value-at-Risk (CVaR), long studied in finance \citep{BO11, Tam15}. Extensions based on state augmentation have also been explored for embedding risk measures directly into MDP formulations \citep{HJ15, Car16, Cho18}. These contributions have found applications in domains such as finance, healthcare, and robotics, where reliability is essential.

Building on the above developments in risk-sensitive, robust, safe, and distributional reinforcement learning, we focus here on the specific challenge of reliable reinforcement learning, and more precisely on maximizing the probability that the total return exceeds a prescribed minimum threshold. To tackle this objective, we propose a state-augmented formulation that embeds the reliability requirement directly into the RL framework by extending the state space with additional information related to the threshold constraint. This approach provides a principled means of aligning reliability objectives with standard RL techniques. We demonstrate its applicability through the problem of reliable routing, where the task is to maximize the probability of reaching a target destination within a given time budget.

\section{Preliminaries }

In this section, we review the fundamental notions of reinforcement learning (RL) and risk-sensitive control that will serve as the foundation of our framework. Our objective is to set the formalism, establish notations, and recall essential definitions.

We consider the standard setting of an agent interacting with a stochastic environment in discrete time steps $t = 0,1,2,\ldots$, with the goal of optimizing a performance criterion. At each step, the agent observes the current state $s_t \in \mathcal S$ of the environment (full observability is assumed) and selects an action $a_t \in \mathcal A$. The system then evolves to a new state $s_{t+1} \in \mathcal S$, and the agent receives a numerical reward $r_{t+1} \in \mathbb R$.

The environment dynamics are modeled as a Markov Decision Process (MDP) with transition probability function
\begin{equation}\label{dyn1}
p(s',r \mid s,a) := \mathbb P (S_{t+1} = s', R_{t+1} = r \mid S_t = s, A_t = a),
\end{equation}
which specifies the probability of reaching next state $s'$ and receiving reward $r$ when taking action $a$ in state $s$.

The expected reward associated with a state–action–next-state triple is defined as
\begin{equation} \nonumber
r(s,a,s') := \mathbb E \big[ R_t \mid S_{t-1} = s, A_{t-1} = a, S_t = s' \big].
\end{equation}

The return from time $t$ is given by the discounted sum of future rewards:
\begin{equation}\nonumber 
G_t := \sum_{k=0}^\infty \gamma^k R_{t+k+1}, \quad \gamma \in [0,1),
\end{equation}
where $\gamma$ is the discount factor.

A policy $\pi(a \mid s)$ defines the probability of selecting action $a$ in state $s$. The classical RL objective is to find a policy that maximizes the expected return:
\begin{equation}\label{optimal1}
\max_{\pi} \mathbb E_{\pi} \Bigg[ \sum_{k=0}^\infty \gamma^k R_{t+k+1} \Bigg],
\end{equation}
subject to the dynamics~\eqref{dyn1}.

A central concept in RL is the action-value function (or $q$-function), which measures the expected return when starting in state $s$, taking action $a$ in $s$, and thereafter following policy $\pi$:
\begin{equation}\nonumber
q_{\pi}(s,a) := \mathbb E_{\pi} \Bigg[ G_t \big| S_t = s, A_t = a \Bigg]
= \mathbb E_{\pi} \Bigg[ \sum_{k=0}^\infty \gamma^k R_{t+k+1} \Big| S_t = s, A_t = a \Bigg].
\end{equation}

The optimal action-value function is obtained by maximizing over all policies:
\begin{equation}\nonumber 
q^*(s,a) := \max_\pi q_\pi(s,a), \qquad \forall s \in \mathcal S, \forall a \in \mathcal A.
\end{equation}

This optimal function satisfies the Bellman optimality equation:
\begin{equation}\label{bellman1}
q^*(s,a) = \sum_{s',r} p(s',r \mid s,a) \Big[ r + \gamma \max_{a'} q^*(s',a') \Big], \qquad \forall s \in \mathcal S, \forall a \in \mathcal A.
\end{equation}

Equation~\eqref{bellman1} lies at the core of dynamic programming and forms the basis of RL algorithms. Since solving it exactly is typically infeasible in large or unknown environments, RL methods approximate its solution using experience collected by the agent.

RL algorithms differ in how they approximate the Bellman equation. Many rely on temporal-difference (TD) learning, which combines dynamic programming ideas with Monte Carlo sampling of trajectories. One of the most fundamental algorithms is Q-learning, which provides an off-policy estimate of $q^*$ and has inspired numerous extensions, including modern deep RL methods that approximate value functions with neural networks.

\subsection*{Risk-sensitive control}

In many decision-making problems, it is not sufficient to optimize only the expected return, since this criterion ignores the variability of outcomes that may arise from uncertainty in the system dynamics or incomplete knowledge of the environment. This variability, commonly referred to as risk, can stem either from intrinsic stochasticity in the process or from epistemic uncertainty about the true state of the world. While risk-neutral control focuses exclusively on maximizing the expectation of the return—corresponding to the objective in~\eqref{optimal1}—such an approach may lead to solutions that perform well on average but are highly unreliable in practice.

To address this limitation, risk-sensitive control explicitly incorporates risk into the decision-making criterion. Instead of optimizing the mean return, the objective is reformulated in terms of a risk measure, which accounts for the distribution of possible outcomes and their associated probabilities. Depending on the chosen risk measure, the controller may favor policies that are more conservative, by penalizing undesirable outcomes (e.g., high variance or catastrophic failures), or more opportunistic, by emphasizing favorable but rare events.

Various risk measures have been studied in the literature, including variance-based criteria, exponential utility functions, and coherent measures such as Value-at-Risk (VaR) and Conditional Value-at-Risk (CVaR). These alternatives enable the design of policies that go beyond average-case performance, striking a balance between reward maximization and robustness to uncertainty. In this sense, risk-sensitive control provides a principled framework for decision-making under uncertainty, complementing the classical risk-neutral paradigm and laying the foundations for reliable reinforcement learning.

Several risk measures are considered in the literature:

\begin{itemize}
  \item[-] Mean Variance criterion: for $\lambda \geq 0$:
     $$\max_{\pi} \mathbb E_{\pi} [G^{\pi}] - \lambda Var (G^{\pi})$$
  \item[-] Entropic risk: for $\lambda \geq 0$:
     $$\max_{\pi} - \frac{1}{\lambda} \ln \mathbb E [e^{-\lambda G^{\pi}}]$$
  \item[-] Value at risk $VaR_{\tau} (G^{\pi}) := F^{-1}_{G^{\pi}} (\tau)$,  where $F_{G^{\pi}} (\rho) := P(G^{\pi} \leq \rho)$ 
              and $F^{-1}_{G^{\pi}} (\tau) := \min \{\rho \geq 0, F_{G^{\pi}} (\rho) \geq \tau \}$. \\
              For $\tau \in (0,1)$, maximize the $\tau$th quantile of the return distribution:
     $$\max_{\pi} F^{-1}_{G^{\pi}} (\tau),$$              
  \item[-] Conditional value at risk $CVaR_{\tau}(G^{\pi}) := \frac{1}{\tau} \int_0^{\tau} F^{-1}_{G^{\pi}}(u) du$.\\
         For $\tau \in (0,1)$, maximize the expected return conditioned on the event that this return is no greater than
         the return's $\tau$th quantile:
     \begin{equation}\label{cvar1}    
       \max_{\pi} \frac{1}{\tau} \int_0^{\tau} F^{-1}_{G^{\pi}}(u) du
     \end{equation}    
     Indeed, when $F^{-1}_{G^{\pi}}$ is strictly increasing, we have 
             $CVaR_{\tau}(G^{\pi}) = \mathbb E [G^{\pi} \mid G^{\pi} \leq F^{-1}_{G^{\pi}}(\tau)]$. \\
     $CVaR_{\tau}$ objective is optimized with a state augmentation procedure by introducing a desired minimum return or 
     target $b \in \mathbb R$ to keep track of the amount of dicounted reward that should be obtained.         
     It was shown that, for $\tau \in (0,1)$, we have \citep{Rockafellar2001}: 
     \begin{equation}\label{cvar2}     
       CVaR_{\tau}(G^{\pi}) = \max_{b \in \mathbb R} \left(b - \tau^{-1} \mathbb E \left[(b - G^{\pi})^+\right] \right),
     \end{equation}  
     where $(x)^+ := \max(x,0)$. \\
     The advantage of this formulation is that it becomes easier to optimize in the context of a policy-dependent return. 
     Indeed (\ref{cvar1}) can now be written:
     \begin{equation}\label{cvar3}
       \max_{\pi} CVaR_{\tau}(G^{\pi}) = \max_{\pi} \max_{b \in \mathbb R} 
                           \left(b - \tau^{-1} \mathbb E \left[(b - G^{\pi})^+\right] \right).
     \end{equation}                 
     The optimization is then done by minimizing the \textit{shortfall} $\mathbb E \left[(b - G^{\pi})^+\right]$ 
     with dynamic programming, and by adjusting $b$ appropriately; see \citep{DRL2023} for example.
   \item[-] Exponential return in average-return MDPs \citep{Arr71,HM72} \\
     An exponential utility function is commonly used to capture risk preferences:
     \begin{equation}
       \max_{\pi} \lim_{T \to \infty} \sup \frac{1}{T} \frac{1}{\beta} \ln \mathbb 
                 E \left[ e^{\beta \sum_{t=0}^{T-1} R_t} \right],
     \end{equation}  
     where $\beta$ is a parameter that controls rsik sensitivity. We notice that:
     \begin{itemize}
        \item[$\bullet$] $R_t > 0$ and $\beta > 0$ corresponds to risk-averse setting.
        \item[$\bullet$] $\beta < 0$ corresponds to risk seeking setting.
        \item[$\bullet$] $\beta \to 0$ corresponds to risk-neutral average return. 
     \end{itemize}
\end{itemize}

\section{Reliable Reinforcement Learning (R2L)}

In this section, we introduce our novel formulation of reliable reinforcement learning (R2L), present the theoretical results, and discuss their consequences and interpretations. Unlike standard reinforcement learning, which optimizes the expected return, our approach explicitly targets the probability of exceeding a given return threshold. This shift in perspective reflects a fundamental change in the optimization criterion: instead of seeking policies that are optimal on average, we focus on policies that are reliable, i.e., those that maximize the likelihood of achieving a minimum guaranteed performance. In doing so, we formalize a new class of reinforcement learning problems where optimality is defined in terms of reliability.

The motivation for this approach arises from the limitations of expectation-based criteria in safety-critical or risk-sensitive contexts. An agent may achieve high expected performance while still facing a significant probability of catastrophic failure, which is unacceptable in applications such as autonomous driving, medical treatment planning, or mission-critical communication networks. By formulating the objective in terms of threshold satisfaction probabilities, reliable RL enables the design of policies that balance efficiency with guarantees of safety and robustness, ensuring that critical performance constraints are met with high confidence. Our theoretical contribution lies in recasting this reliability-oriented objective into a tractable RL framework through a state-augmented representation, which will be developed in detail below.

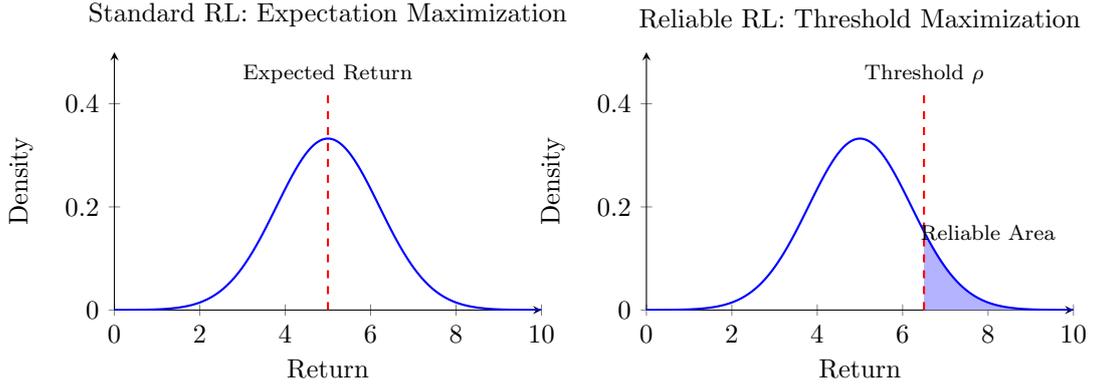
\begin{figure}[htbp]
\centering
\begin{tikzpicture}

\begin{axis}[
    width=0.45\textwidth,
    height=5cm,
    domain=0:10,
    samples=200,
    axis lines=left,
    xlabel={Return},
    ylabel={Density},
    title={Standard RL: Expectation Maximization},
    ymin=0, ymax=0.5,
    xtick={},
    ytick={}
]

\addplot[blue, thick] {1/(1.2*sqrt(2*pi)) * exp(-(x-5)^2/(2*1.2^2))};

\draw[red, thick, dashed] (axis cs:5,0) -- (axis cs:5,0.42) node[above, black] {\footnotesize Expected Return};

\end{axis}



\begin{axis}[
    at={(7cm,0cm)},
    width=0.45\textwidth,
    height=5cm,
    domain=0:10,
    samples=200,
    axis lines=left,
    xlabel={Return},
    ylabel={Density},
    title={Reliable RL: Threshold Maximization},
    ymin=0, ymax=0.5,
    xtick={},
    ytick={}
]

\addplot[name path=A, blue, thick] {1/(1.2*sqrt(2*pi)) * exp(-(x-5)^2/(2*1.2^2))};

\draw[red, thick, dashed] (axis cs:6.5,0) -- (axis cs:6.5,0.42) node[above, black] {\footnotesize Threshold $\rho$};

\path[name path=B] (axis cs:6.5,0) -- (axis cs:10,0);
\addplot[blue!30] fill between[of=A and B, soft clip={domain=6.5:10}];

\node[black] at (axis cs:8,0.15) {\footnotesize Reliable Area};

\end{axis}

\end{tikzpicture}

\caption{Comparison between standard RL (maximizing expected return) and reliable RL (maximizing the probability of exceeding a threshold).}
\label{fig:reliableRL}
\end{figure}

The relevance of reliable strategies can be illustrated through several motivating applications. A first example is reliable routing in transportation or communication networks. Rather than minimizing the expected travel or transmission time between an origin–destination pair, the goal is to maximize the probability of arriving within a fixed time budget (see, e.g., \citep{FN06}, \citep{SBB11}). In this context, the agent seeks the most reliable route, defined as the one that guarantees arrival within the specified time threshold. This example will be treated in detail in Section~\ref{sec-example}, where we demonstrate the practical application of our framework to the Stochastic On-Time Arrival problem.

A second example arises in sequential games with multiple stages, where advancing to the next stage requires achieving a minimum number of points or tokens. The agent must adopt policies that maximize the probability of meeting or surpassing the threshold score at each stage, thereby ensuring progression. The notion of reliability thus translates into the guarantee of reaching the required performance level consistently, rather than merely maximizing expected points.

A third example can be found in multi-agent communications and information routing, where the goal is to ensure that information reaches a given agent—or all agents in a network—within a prescribed time bound. Analogous to reliable routing, the objective is not to minimize the average communication delay but to maximize the probability that delivery occurs before the deadline. Reliable RL thus provides a principled framework for deriving strategies that guarantee dependable communication under stochastic and adversarial conditions.

Through these applications, we see that reliability-oriented optimization is not merely a theoretical construct but a practical necessity in many domains. In the remainder of this section, we formalize this problem, derive its theoretical underpinnings, and demonstrate how our state-augmented formulation enables its resolution within the standard RL paradigm.

\subsection*{Maximize the probability of exceeding a return threshold}

In order to formalize this idea of optimizing the probability of exceeding a given return threshold, instead of maximizing the expectation of the return, let us consider the following function:

$q_{\pi}(s,a,\rho) :=$ probability to exceed return $\rho$, starting with state $s$, and taking action $a$ at $s$, under policy $\pi$:
\begin{equation}\label{def_u}
  q_{\pi}(s,a,\rho) := \mathbb P_{\pi}\left(G_t \geq \rho \mid S_t = s, A_t = a\right)
\end{equation}  

Then we consider the problem of optimizing the probability $q_{\pi}(s,a,\rho)$:
\begin{equation}\label{optimal2}
  \max_{\pi} q_{\pi}(s,a,\rho),
\end{equation}
under the dynamics (\ref{dyn1}).

Let us now denote $q^*(s,a, \rho) := \max_{\pi} q_{\pi}(s,a,\rho)$. Then we have:
\begin{align}
q^*(s,a,\rho) & = \max_{\pi} q_{\pi}(s,a,\rho) \nonumber \\
              & = \max_{\pi} \mathbb P_{\pi}\left( G_t \geq \rho \mid S_t = s, A_t = a \right) \nonumber \\
              & = \max_{\pi} \mathbb P_{\pi} \left( \sum_{k=0}^{+\infty} \gamma^k R_{t+k+1} \geq \rho \mid S_t = s, A_t = a\right) \nonumber \\
              & = \gamma \int_{s'} \int_0^{\rho} p(s',r \mid s,a) \max_{a'}q^*(s',a',\rho - r) \; dr\;ds'. \qedhere \label{bellman2}
\end{align} 

Equation (\ref{bellman2}) is written for $\rho \in \mathbb R$. However, it is often more practice to limit $\rho$ in a given interval $[\underline{\rho}, \bar{\rho}]$.
For example, if we know that the rewards $R_t$ are bounded: $\underline{R}_t \leq R_t \leq \bar{R}_t, \forall t \geq 0$, then, we deduce that
$\underline{R}_t /(1-\gamma) \leq G_t \leq \bar{R}_t/(1-\gamma)$. 
In this case, we can take $\underline{\rho} = \underline{R}_t /(1-\gamma)$ and $\bar{\rho} = \bar{R}_t/(1-\gamma)$.
Therefore, we have also the following equations:
\begin{align}
  & q^*(s,a,\rho) = 1, \forall s\in\mathcal S, a\in\mathcal A, \rho \leq \underline{\rho}, \label{eqdown} \\
  & q^*(s,a,\rho) = 0, \forall s\in\mathcal S, a\in\mathcal A, \rho > \bar{\rho}. \label{eqtop}
\end{align}
In this case, to avoid out of bounds values of $(\rho - r)$ in Equation (\ref{bellman2}), we can replace $(\rho - r)$ with 
$\max (\underline{\rho}, \min (\bar{\rho}, \rho - r))$. 

For example, if we know that all the rewards are non negative $R_t \geq 0, \forall t\geq 0$, then we have $\underline{\rho} = 0$ and Equation (\ref{bellman2}) is 
completed with the following equation:
\begin{equation}\label{bellman22}
  q^*(s,a,\rho) = 1, \forall s\in\mathcal S, a\in\mathcal A, \rho \leq 0.
\end{equation}
In this case, to avoid negative values of $(\rho - r)$ in Equation (\ref{bellman2}), we can replace $(\rho - r)$ with 
$\max (0, \rho - r)$. 

Let us now consider the following optimization problem (\ref{p1}) in standard formulation, defined by: the set of states, set of actions, set of rewards, transition probability distribution, set of terminal states, criterion, and Bellman equation, respectively:
\begin{equation}\label{p1}
   (\mathcal S, \mathcal A, \mathbb R, p, \emptyset, (\ref{optimal2}), (\ref{bellman2})-(\ref{eqtop}))  \tag{P1}
\end{equation}  

We present in the following subsection the state-augmented formulation proposed in this work. This formulation consists in extending the state representation to include, in addition to the standard state $s$, the remaining return $\rho$ that must still be exceeded starting from $s$, relative to an initially predefined total return threshold. By introducing this augmented state space, the reliable reinforcement learning problem is reformulated as a standard optimization problem that remains solvable using conventional reinforcement learning algorithms.

\subsection{State augmented formulation}

We propose in this section a new formulation 
where:
\begin{itemize}
  \item The observed state is: $\sigma := (s,\rho)$ : the pair composed of the state $s$ in standard 
    formulation, and the remaining return $\rho$ that must still be exceeded starting from $s$,  
    relative to an initially predefined total return threshold. 
  \item The action is: $a$: the same as in the standard formulation.
\end{itemize}

Maximizing the probability that the return $G_t$ exceeds a given threshold is equivalent to maximizing the probability that the cumulative return accumulated along the transitions reaches this predefined threshold. In the proposed state-augmented formulation, this objective can be reformulated as maximizing the probability of reaching any augmented state $\sigma = (s, \rho)$ such that $s \in \mathcal{S}$ and $\rho \leq \underline{\rho}$.
For instance, in the particular case where $\underline{\rho} = 0$ (corresponding to non-negative rewards), maximizing the probability that the return $G_t$ exceeds a given threshold becomes equivalent to maximizing the probability of reaching any augmented state $\sigma = (s, 0)$—that is, the set of states where the remaining return to exceed is zero, meaning that the initially fixed total return threshold has already been achieved. 

The objective is then written in the state augmented formulation as follows:
\begin{equation}
   \max \mathbb P\left( \text{reach any state } \sigma = (s, \rho) \right), \text{with }\rho \leq \underline{\rho}.
\end{equation}

Let us denote $\tilde{q}_{\pi} (\sigma,a) := q_{\pi}(s,a,\rho)$. Then we have:
\begin{align}
   \tilde{q}_{\pi} (\sigma,a) & = \mathbb P_{\pi}\left(G_t \geq \rho \mid S_t = s, A_t = a\right) \\
                              & = \mathbb P_{\pi} \left( \text{reach } (s', \rho'), \text{ with } \rho' \leq \underline{\rho} \mid \sigma_t=(s,\rho), A_t = a\right).
\end{align}   
We then denote by $\tilde{q}^*(\sigma,a) := q^*(s,a,\rho)$ the optimal action-value function in the state-augmented formulation. Then we have:
\begin{align}
\tilde{q}^*(\sigma,a) & = \gamma \int_{s'} \int_0^{\rho} p(s',r \mid s,a) \max_{a'} 
                            q^*(s',a',\rho-r) \; dr \; ds'. \nonumber\\
                      & = \gamma \int_{\sigma'} p(\sigma' \mid \sigma, a) \max_{a'} 
                            \tilde{q}^*(\sigma',a') \; d\sigma' \label{eq:qtilde}
\end{align}              
with
\begin{equation}\label{eq:terminal}
  \tilde{q}^*(\sigma,a) = 1, \forall \sigma=(s,\rho) \text{ with } \rho \leq \underline{\rho}.
\end{equation}  


Let us now denote $\sigma^*$ any terminal state $(s,\rho)$ with $\rho \leq \underline{\rho}$. 
We then consider the following optimization problem in the state-augmented formulation:
\begin{equation}\label{pbm2}
   \max_{\pi} \mathbb E_{\pi} \sum_{\sigma \neq \sigma^*} \tilde{R}_{\sigma}^a + \tilde{R}_{\sigma^*},
\end{equation}
where the reward in the state-augmented formulation is defined:
\begin{equation}
    \begin{cases}
       \tilde{R}_{\sigma}^a := 0 & \text{ for non terminal } \sigma, \text{ i.e. } \forall s \in \mathcal S, 
          \forall \rho > \underline{\rho} , \\
       \tilde{R}_{\sigma^*} := 1 & \text{ for terminal } \sigma = \sigma^*, \text{ i.e. } \forall s \in \mathcal S, 
          \forall \rho \leq \underline{\rho},
    \end{cases} \label{r2} 
\end{equation}
and where the dynamics of the system are given by the probability distribution:
\begin{equation}\label{dyn2}
  p(\sigma'=(s',\rho') \mid \sigma=(s,\rho), a) := p(s', \rho-\rho' \mid s,a).
\end{equation} 

\begin{proposition}
  In the state-augmented formulation, the Bellman equation of the optimization problem (\ref{pbm2}) under the dynamics (\ref{dyn2}),   
  is given by Eq. (\ref{eq:qtilde})-(\ref{eq:terminal}). 
\end{proposition}
\proof 
From the definition (\ref{r2}) of the reward $\tilde{r}$ in the state-augmented formulation, it is easy to chack that (\ref{eq:qtilde})-(\ref{eq:terminal}) can be written:
\begin{align}
  & \tilde{q}^*(\sigma,a) = \int_{\sigma'} p(\sigma' \mid \sigma, a) [\tilde{r} + \gamma \max_{a'} \tilde{q}^*(\sigma',a')] d\sigma',
     \quad \forall \sigma \neq \sigma^*, \label{bell21} \\
  & \tilde{q}^*(\sigma^*,a) = 1.  \label{bell22}
\end{align} 
which gives the result. 
\endproof

Let us now consider the following optimization problem (\ref{p2}) in state augmented formulation, defined by the set of states, set of actions, set of rewards, transition probability distribution, set of terminal states, criterion, and Bellman equation, respectively:
\begin{equation}\label{p2}
  (\mathcal S \times \mathbb R, \mathcal A, \{0,1\}, (\ref{dyn2}), \{(s, \rho), \rho \leq \underline{\rho}\}, (\ref{pbm2}), (\ref{bell21})-(\ref{bell22}))  \tag{P2}
\end{equation}  

\noindent
Then we have the following result.

\begin{proposition}\label{prop2}
   Solving the optimization problem~(\ref{p1}) in the standard formulation is equivalent to solving the optimization problem~(\ref{p2}) in the state-augmented formulation.
\end{proposition}

\proof
The quivalence of the two problems (\ref{p1}) and (\ref{p2}) comes from the equivalence of the two Belman equations (\ref{bellman2})-(\ref{bellman22}) and (\ref{bell21})-(\ref{bell22}).
\endproof

The main consequence of \newref{Proposition}{prop2} is that any existing RL or DRL algorithm can be directly applied to solve the optimization problem (\ref{p2}) in its state-augmented formulation. In doing so, one simultaneously addresses the original optimization problem (\ref{p1}) in the standard formulation. This result effectively extends the classical reinforcement learning paradigm, where the objective is to maximize the expected return, to a reliability-oriented framework in which the goal is to maximize the probability of exceeding a prescribed return threshold. The policies obtained under this formulation can thus be regarded as reliable strategies, offering performance guarantees that go beyond expectation optimality. Moreover, the framework allows the agent to balance two complementary objectives: achieving high expected returns on average and ensuring a sufficiently high probability of exceeding a minimum guaranteed return threshold. This trade-off is particularly valuable in safety-critical or uncertainty-sensitive applications, such as reliable routing under time budgets, sequential games requiring stage completion, or multi-agent communication systems with strict delivery deadlines, where reliability often outweighs purely average efficiency.

We explicit below in Algorithm~\ref{al1} the Q-learning algorithm to be used for solving problem (\ref{p2}):

\begin{algorithm}[H]
\caption{\textbf{\; - \; Reliable Q-learning algorithm}}
\label{al1}
\begin{algorithmic}
    \STATE \textbf{Input: } Discount factor $\gamma$,  Learning rate $\alpha$, Return lower bound $\underline{\rho}$, 
      Return upper bound $\bar{\rho}$
    \STATE Initialize $Q(s,a, \rho), \forall s\in \mathcal S, \forall a \in \mathcal A(s), 
         \forall \rho \in (\underline{\rho}, \bar{\rho})$, 
         and fix $Q(\cdot,\cdot,\underline{\rho}) = 1$ 
    \BlankLine
    \FOR{ each episod}
      \STATE Initialize $s$ and $\rho$ 
      \BlankLine
      \FOR{ each step of episod}
                \STATE Choose $a$ from $(s, \rho)$ using a policy derived from $Q$ (e.g. $\varepsilon$-greedy) 
                \STATE Take action $a$, observe $r$, $s'$ 
                  and put $\rho' := \max\left(\underline{\rho}, \min(\bar{\rho}, \rho-r)\right)$  
                \STATE $Q(s,a,\rho) := Q(s,a,\rho) + \alpha [\gamma \max_{a'} Q(s',a', \rho') - Q(s,a,\rho)]$                
                \STATE $s := s'$ and $\rho := \rho'$ 
      \ENDFOR{ Until $\rho = \underline{\rho}$ (terminal state) Or Maximum number of steps reached }  
      \BlankLine
    \ENDFOR{ Until convergence }    
\end{algorithmic}
\end{algorithm}

We presented in Algorithm~\ref{al1} the reliable extension of the well-known Q-learning algorithm. As established in Proposition~\ref{prop2}, the equivalence between problems (\ref{p1}) and (\ref{p2}) ensures that any existing reinforcement learning algorithm—whether classical or deep—can be adapted to maximize the probability of exceeding a prescribed return threshold and thereby derive reliable strategies. In practice, this means that one can directly apply an existing RL algorithm to the state-augmented formulation (\ref{p2}), and by doing so, also obtain a solution to the original problem (\ref{p1}). This observation highlights the generality and flexibility of our framework: it extends the applicability of a broad class of RL methods from expectation-based objectives to reliability-oriented objectives. In the next section, we demonstrate the effectiveness of this approach through an application to reliable routing. Specifically, we implement our framework using the Dueling Double DQN algorithm \citep{Has16,Wang16}, a state-of-the-art deep RL method that combines the stability of Double DQN with the representational advantages of dueling network architectures. The reliable adaptation of this algorithm, which we denote Reliable Dueling Double DQN (RD3QN), is detailed in Algorithm~\ref{alg:d3qn} of Appendix~\ref{RD3QN}.

\subsection{Price of Reliability}
\label{sec-por}

We now provide an interpretation of the Q-functions $q^*(s,a,\rho)$ obtained from the reliable reinforcement learning formulation. These optimal Q-values, expressed as functions of the remaining return $\rho$, can be viewed as complementary cumulative distribution functions (CCDFs). Indeed, they represent the probability of exceeding a given reward threshold, which is complementary to the classical cumulative distribution function (CDF) that gives the probability of not exceeding that threshold. Consequently, while a CDF is non-decreasing with respect to the variable of interest, the functions $q^*(s,a,\rho)$ are by contrast non-increasing with respect to $\rho$.

\begin{figure}[h]
        \centering
        \includegraphics[height=7cm]{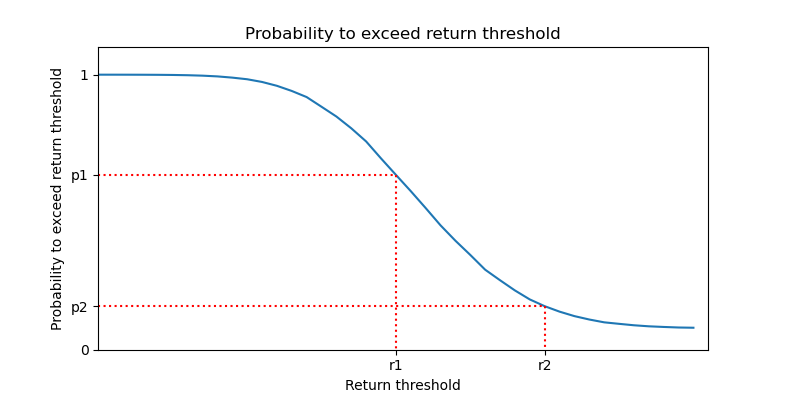}
        \caption{The price of reliability.}
        \label{fig:por0}
\end{figure} 

Figure~\ref{fig:por0} illustrates this interpretation by representing the maximum probability $q^*(s,a,\rho)$ as a function of the reward threshold $\rho$. As shown in the figure, the probability of exceeding a reward threshold $r_1$ corresponds to $p_1$, whereas the probability of exceeding a higher threshold $r_2$, with $r_2 > r_1$, is given by $p_2$, where $p_2<p_1$. This monotonic decrease reflects the fundamental trade-off between performance and reliability: higher reward thresholds are increasingly difficult to achieve, and thus correspond to lower probabilities of success.

The curve also provides a direct means of quantifying the reliability–performance trade-off. The \textit{price} of moving from a threshold $r_1$ to a higher threshold $r_2$ is a reduction in reliability from $p_1$ to $p_2$. In other words, at the level $r_1$, the gain of $(r_2 - r_1)$ in expected reward comes at a cost of $(p_1 - p_2)$ in reliability. This marginal cost depends on the local slope of the curve and may vary across different values of $\rho$.

Conversely, one can interpret the figure from the reliability perspective: to increase the reliability from $p_2$ to $p_1$, one must accept a reduction in the attainable reward threshold from $r_2$ to $r_1$. Thus, at the reliability level $p_2$, the gain 
$(p_1 - p_2)$ in reliability corresponds to a loss of $(r_2 - r_1)$ in achievable reward. This bidirectional interpretation highlights the fundamental trade-off surface between reliability and reward, which plays a central role in designing risk-sensitive or reliability-aware reinforcement learning policies.

\section{Illustrative example: Reliable routing}
\label{sec-example}

To illustrate the results presented above, we consider the application of our framework to reliable routing. In classical routing optimization, the objective is typically to minimize the expected travel time between a given origin and destination in a network. Reliable routing, by contrast, seeks to maximize the probability of reaching the destination within a prescribed time budget, starting from a given origin. A key advantage of this formulation is that it allows the determination of an optimal itinerary tailored to each time budget, thereby maximizing the likelihood of on-time arrival. Such itineraries are referred to as reliable because they provide a probabilistic guarantee—namely, the maximum probability of reaching the destination within the specified time. This problem is widely known in the literature as the Stochastic On-Time Arrival (SOTA) problem.

The SOTA problem has been extensively studied in the operations research and transportation literature. \citep{FN06} first formulated it as a stochastic dynamic programming problem and solved it using successive approximation. Building on this, \citep{NF06} proposed a discrete approximation algorithm with finite-step convergence and pseudo-polynomial complexity. Exact solutions were later established in \citep{SBB11}, even in cyclic networks, under the assumption of minimum realizable link travel times. A major computational challenge of these methods lies in the repeated evaluation of continuous-time convolution products. To mitigate this cost, speedup and pruning techniques were introduced in \citep{Sam12, KSS14}. More recently, computational efficiency has been improved by parallelization strategies: \citep{AS24} developed a CUDA-based GPU implementation that achieves orders-of-magnitude speed-ups compared to traditional single-threaded CPU solvers. The SOTA problem has also been extended to robust routing \citep{AIP15, JTTE17, TRP17, EJOR20}.
Reliable routing naturally aligns with the concept of reliable reinforcement learning, since the objective of maximizing the probability of on-time arrival can be directly formulated as maximizing the probability of exceeding a threshold on cumulative return.

We consider the problem of reliable routing in a given network and from a given origin node to a given destination node. In this context, reliable routing refers to the design of policies that maximize the probability of reaching the destination node within a prescribed time budget. This reliability-oriented objective differs fundamentally from the classical shortest-path or routing formulations, which focus on minimizing the expected travel time between two nodes. In reliable routing, the central concern is not the average-case efficiency but the guarantee of timely arrival with high probability. Consequently, an optimal reliable policy may deliberately accept a longer expected travel time if doing so increases the likelihood of meeting the time budget. This formulation naturally integrates with our state-augmented reinforcement learning framework, where the remaining time budget is treated as part of the state, thereby allowing standard RL algorithms to derive policies that explicitly optimize for reliability rather than mere expectation.

\subsection{Formulation of the reliable routing problem}

We consider a graph with finite number of nodes and edges (links).
Given a node $i \in \mathcal{I}$ and a time budget $t$, let $v_i(t)$ denote the probability of successfully reaching a prefixed destination node $d$ from node $i$ within a prescribed time budget $t$. In this routing problem, the observable state on which both actions and policies are conditioned is the current node $i \in \mathcal{I}$. A routing action, denoted by $a_i(t)$, specifies the choice of the subsequent node when located at node $i$ with a remaining time budget $t$. We denote by $v^*_i(t)$ the \emph{maximum} probability of reaching the destination node $d$ from node $i$ within the time budget $t$, i.e., the probability obtained under the optimal routing policy. The functions $v_i(t)$ and $v^*_i(t)$ therefore play the roles of the value function and the optimal value function, respectively, in this reliability-oriented routing problem.  

For each node $i \in \mathcal{I}$, let $\Gamma(i)$ denote the set of its immediate successor nodes in the network. The stochastic travel times associated with traversing a link $(i,j)$ are characterized by probability density functions (pdfs) $p_{ij}(\cdot)$. Throughout this work, we consider the standard case where link travel times are independent and not temporally or spatially correlated. The pdfs of link travel times are assumed to be known, and may be obtained, for example, from historical traffic data, empirical measurements, or real-time traffic information. This modeling assumption ensures that the optimization framework remains both tractable and practically relevant, while providing a probabilistic foundation for the derivation of reliable routing strategies.  

Moreover, this formulation naturally leads to a recursive characterization of the optimal value function. Specifically, the probability of reaching the destination within a given time budget from node $i$ can be expressed in terms of the probabilities associated with its successor nodes in $\Gamma(i)$, weighted by the distributions of travel times on the outgoing links. This recursion takes the form of a Bellman-type equation, which constitutes both the theoretical foundation for analyzing reliable routing and the computational basis for applying reinforcement learning algorithms in this setting.  

The maximum probability
$v^*_i\left(t\right)$ (optimal value function) and the optimal successor node $a^*_i\left(t\right)$ (optimal policy) satisfy the following equations:
\begin{align}
  & v^*_i \left( t \right) = \max_{j \in \Gamma (i)} \int_0^t p_{\mathit{ij}}\left(w\right)v^*_j\left(t-w\right)\mathit{dw}, \qquad \forall i \in \mathcal I \setminus\{d\}, 0{\leq}t{\leq}T  \label{sota1}\\
  & v^*_d\left(t\right) = 1, \qquad 0{\leq}t{\leq}T \label{sota2}\\
  & a^*_i\left(t\right) = \mathit{arg} \max_{j \in \Gamma (i)} \int_0^t p_{\mathit{ij}}\left(w\right)v^*_j\left(t-w\right)\mathit{dw}, \qquad \forall i \in \mathcal I \setminus \{d\}, 0{\leq}t{\leq}T \label{sota3}
\end{align}
where $T$ is the maximum time budget.

It is important to note that in this routing problem the transition to the next state is deterministic once an action has been taken. More precisely, the action at node $i$ consists of selecting one of its immediate successor nodes $j \in \Gamma(i)$. Since the action directly specifies the next node to be visited, the next state coincides with the chosen action. In other words, the mapping between actions and successor states is one-to-one: once the user selects an action, i.e., one immediate successor of node $i$, this choice uniquely determines the subsequent state of the environment $s_{t+1} = a_t$. This property simplifies the state-transition dynamics compared to more general Markov decision processes, where the next state is typically stochastic given an action, and highlights that the source of uncertainty in this problem arises solely from the stochasticity of the travel times on the links rather than from the state-transition structure itself.  

We now reformulate the reliable routing problem in the framework of Problem~(\ref{p1}).  
To achieve this, the recursive equations (\ref{sota1})--(\ref{sota2}) must be expressed in the form of the generalized Bellman equations (\ref{bellman2})--(\ref{bellman22}).  
In particular, we introduce the action-value function $q^*(i,j,t)$, which represents the maximum probability of reaching the destination node $d$ within a remaining time budget $t$, starting from node $i$ and choosing as the first action the immediate successor node $j \in \Gamma(i)$.  
Then from (\ref{sota1})-(\ref{sota2}) we can write:
\begin{align}
  & q^*(i,j,t) = \int_0^t p(w\mid i,j) \max_{k \in \Gamma (j)} q^*(j,k,t-w) \; dw,  \qquad \forall i \in \mathcal I \setminus\{d\}, 0{\leq}t{\leq}T  \label{sota11}\\
  & q^*(d,j,t) = 1, \qquad \forall j\in \Gamma (d), \forall t, 0{\leq}t{\leq}T. \label{sota12}
\end{align}
In (\ref{sota11}) we do not have an integral on the next state $s'$, as in (\ref{bellman2}), because, as mentioned above, in the routing problem, once the action $j$ (immediate successor node) is taken, the next state is given deterministically, and coincides with $j$.

Let us now write the problem in the framework of Problem (\ref{p2}). For that, let $\sigma := (i,t)$ denote the state of the environment in the state-augmented formulation, composed of the current node $i$ and the remaining time budget $t$; and let $\tilde{q}^* (\sigma, j)$ be defined as $\tilde{q}^* (\sigma, j) := q^*(i,j,t)$. 
Finally, we define the reward $\tilde{r}$ as follows:
\begin{equation}
   \tilde{r}(\sigma, j) = \begin{cases}
                       0 & \text{ for non terminal } \sigma, \text{ i.e. } \forall i\neq d, \forall 0{\leq}t{\leq}T , \\
                       1 & \text{ for terminal } \sigma, \text{ i.e. } i = d, 0{\leq}t{\leq}T.
                    \end{cases} \label{r3} 
\end{equation}
Then, the Bellman equation in the state-augmented formulation is written as follows: 
\begin{align}
  \tilde{q}^*(\sigma,j) & = \int_{\sigma'} p(\sigma' \mid \sigma, j) [\tilde{r}(\sigma,j) + \max_{k\in \Gamma(j)} \tilde{q}^*(\sigma',k)] \; d\sigma', \qquad
     \forall \sigma=(i,t), i\neq d, 0{\leq}t{\leq}T, \label{bell31} \\
  \tilde{q}^*(\sigma,j) & = 1, \qquad \forall \sigma=(d,t) \text{ with } 0{\leq}t{\leq}T.  \label{bell32}
\end{align}  
We notice that the integral on $\sigma'$ in (\ref{bell31}) integrates only on $t$, since the immediate successor node is fixed to $j$ (which is also the action here).
Therefore, (\ref{bell31})-(\ref{bell32}) can be simplified to:
 \begin{align}
  \tilde{q}^*\left((i,t),j\right) & = \int_0^t p(w \mid i,j) [\tilde{r}\left((i,t),j\right) + \max_{k\in \Gamma(j)} \tilde{q}^*\left((j,t-w),k\right)] \; dw, \qquad
     \forall i\neq d, 0{\leq}t{\leq}T, \label{bell41} \\
  \tilde{q}^*\left((d,t),j\right) & = 1, \qquad \forall 0{\leq}t{\leq}T.  \label{bell42}
\end{align}

The dynamic programming equations~(\ref{bell41})--(\ref{bell42}) provide the foundation for solving the reliable routing problem within the reinforcement learning framework. In particular, these equations establish the recursive structure required to compute the maximum probability of reaching the destination within a prescribed time budget, thereby allowing the problem to be addressed using standard RL techniques once reformulated in the state-augmented setting. Consequently, the reliable routing problem can be solved by adapting existing RL algorithms to this probabilistic objective. In the subsequent subsection, we demonstrate this approach by applying the Reliable Q-learning algorithm (Algorithm~\ref{al1}) to low-dimensional grid networks, where the problem remains computationally tractable. To address larger-scale instances with higher dimensionality and more complex dynamics, we further employ a deep reinforcement learning method, namely the Reliable Dueling Double DQN (given in Appendix~\ref{RD3QN}), which leverages neural function approximation to efficiently learn reliable policies in high-dimensional state spaces. This two-level experimentation highlights both the conceptual clarity of the proposed formulation and its scalability when combined with modern DRL architectures.  

\subsection{Solving the reliable routing problem with R2L}

Let us consider a grid network of dimension n×m with bidirectional arcs, as illustrated in Figure~\ref{grid}, where for concreteness a 5×5 grid is depicted. In the numerical experminents we did in this work, the travel time on each link $(ij)$ follows a Gamma probability distribution with mean $\tau^{mean}_{ij}$ and standard deviation 
$\tau^{sd}_{ij}$. Mean link travel times $\tau^{mean}_{ij}$ are randomly generated in the interval (1, 5) for each link. Standard deviation link travel times $\tau^{sd}_{ij}$ are randomly generated in the interval (0.1, 0.5) for each link.

\subsubsection{Reliable Q-learning algorithm for low-dimensional grid networks}

\begin{figure}
  \centering
  \includegraphics[scale=0.4]{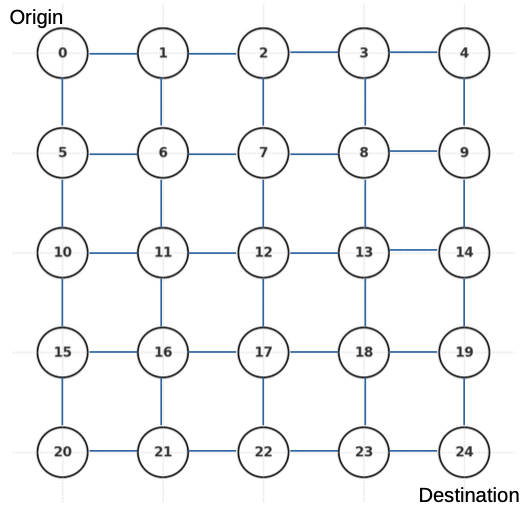}
  \caption{A non directed 5x5 grid network with orgin node 0 and destination node 24.}
  \label{grid}
\end{figure}

In order to better illustrate the approach of Reliable RL for reliable routing, we start by applying the Reliable Q-learning algorithm (Algorithm~\ref{al1}) to low-dimensional grid networks, where the problem remains computationally tractable.
More precisely we consider the non-directed $5\times 5$ grid network of Figure~\ref{grid}.

We highlight here several important specifications of Algorithm~\ref{al2}, designed for the reliable routing problem, in comparison with Algorithm~\ref{al1}, which addresses the general reliable RL formulation. The first key distinction lies in the optimization objective: whereas Algorithm~\ref{al1} maximizes the probability of exceeding a prescribed return threshold (i.e., maximizing cumulative reward), Algorithm~\ref{al2} instead maximizes the probability of arriving at the destination node within a given total time budget (i.e., ensuring that travel time does not exceed the prescribed bound). A second difference concerns the definition of terminal states. In Algorithm~\ref{al1}, a terminal state occurs only when the return threshold $\rho = \underline{\rho}$ is attained, corresponding to the satisfaction of the reliability criterion. In contrast, Algorithm~\ref{al2} involves two types of terminal states: (i) when the time budget is exhausted without reaching the destination, and (ii) when the destination node is successfully reached within the allocated budget. These differences also influence the initialization of the action-value function $Q$. In Algorithm~\ref{al1}, initialization is based on the fact that the probability of exceeding the lower return threshold $\underline{\rho}$ is $1$. For Algorithm~\ref{al2}, initialization reflects routing-specific conditions: starting from any non-terminal node with zero remaining time, the probability of reaching the destination is $0$, whereas starting from the destination node itself, the probability of being at the destination within any nonnegative remaining time is $1$. Finally, the stopping condition for episodes also differs. In Algorithm~\ref{al1}, an episode terminates once the return threshold is achieved or if a prefixed maximum number of steps is reached, while in Algorithm~\ref{al2}, termination occurs either when the destination node is reached or when the time budget has been fully consumed. These distinctions highlight how the general reliable RL framework is adapted to the specific structural and probabilistic features of the reliable routing problem.  
Table \ref{tab1} gives the values fixed for different parameters of Algorithm \ref{al2} for reliable routing.

\begin{algorithm}[H]
\caption{\textbf{\; - \; Reliable Q-learning algorithm for reliable routing}} 
\label{al2}
\begin{algorithmic}
    \STATE \textbf{Input: } Discount factor $\gamma$,  Learning rate $\alpha$, Maximum time budget $T$
    \STATE Initialize $Q(i,j, t) \forall i\in \mathcal I, j \in \Gamma^+(i), \forall t \in [0,T]$,  
    and fix $Q(i \neq d,\cdot,0) = 0$ and $Q(d,\cdot,t) = 1, \forall t \in [0,T]$
    \BlankLine
    \FOR{ each episod}
      \STATE Initialize $i\in\mathcal I$ and $t \in [0,T]$
      \BlankLine      
      \FOR{ each step of episod}
                \STATE Choose $j\in\Gamma^+(i)$ using a policy derived from $Q$ (e.g. $\varepsilon$-greedy)
                \STATE Take successor node (action) $j$, observe travel time $\tau_{ij}$, 
                  and put $t' := \max\left(0, \min(T, t-\tau_{ij})\right)$
                \STATE $Q(i,j,t) := Q(i,j,t)
                         + \alpha [\gamma \max_{k\in\Gamma (j)} Q(j,k, t') - Q(i,j,t)]$               
                \STATE $i := j$ and $t := t'$
      \ENDFOR{~Until $i = d$ or $t = 0$ (end of episod)}           
      \BlankLine 
    \ENDFOR{~Until convergence}    
\end{algorithmic}
\end{algorithm}
 
\begin{table}
  \centering 
  \caption{Reliable Q-learning parameters.}
  \begin{tabular}{rl}
     \hline
     Grid dimension & $5\times 5$ \\
     Minimum remainging time budget & 0 \\
     Maximum remainging time budget $T$ & 30 \\
     Penality for taking forbiden actions (non-existent successor nodes) & 100 \\
     Learning rate $\alpha$ & 1e-04 \\
     Discount rate $\gamma$ & 0.99 \\
     First value of $\varepsilon$ ($\varepsilon$-greedy) & 1 \\
     Epsilon decay & Linear \\
     Number of episodes & 20 M \\
     Maximum number of steps for each episod & 30 \\
     \hline
  \end{tabular}
  \label{tab1}
\end{table}

\begin{figure*}
    \centering
    \begin{subfigure}[b]{1\textwidth}
        \centering
        \includegraphics[height=7cm]{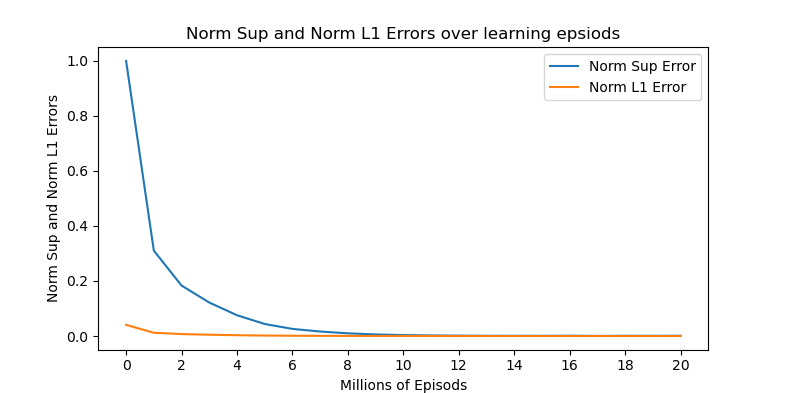}
        \caption{Norm sup and Norm L1 errors over learning episods.}
        \label{fig:errors}
    \end{subfigure} \\
    ~ \hfill
    \begin{subfigure}[b]{1\textwidth}
        \centering
        \includegraphics[height=7cm]{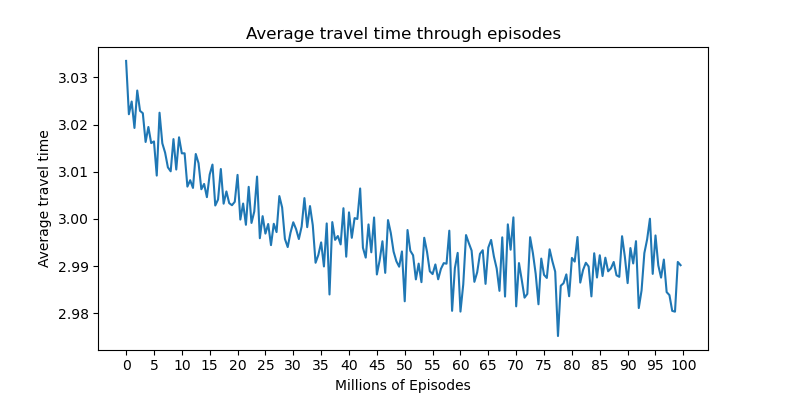}
        \caption{Average link travel time through episods.}
        \label{fig:tt}
    \end{subfigure} \\
     ~ \hfill
    \begin{subfigure}[b]{1\textwidth}
        \centering
        \includegraphics[height=7cm]{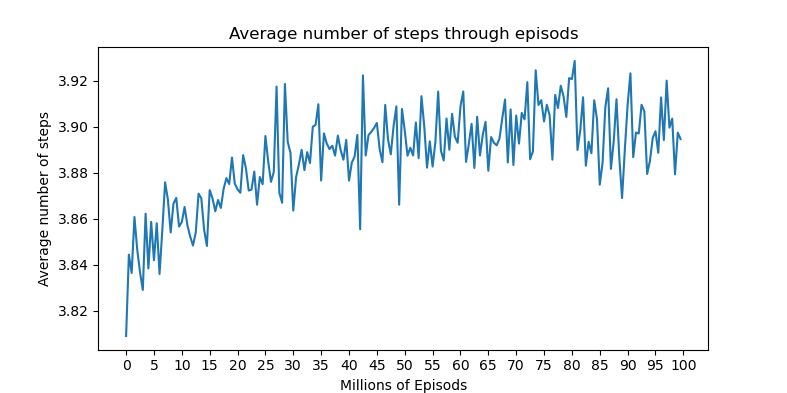}
        \caption{Average number of steps through episods.}
        \label{fig:steps}
    \end{subfigure} \\
    \caption{Illustration of the convergence of the reliable Q-learning algorithm for reliable routing.}
\end{figure*}

Figure~\ref{fig:errors} reports the evolution of the approximation errors on the action value function, measured in terms of the supremum norm and the 
L1 norm, over the course of learning. Both error metrics exhibit a clear monotonic decreasing trend, converging towards zero as the number of episodes increases. This behavior demonstrates the stability and consistency of the proposed reliable Q-learning algorithm when applied to the routing problem. In particular, the initially large supremum norm error decreases rapidly within the first few million episodes, after which it continues to diminish more gradually until convergence. Similarly, the L1 error drops sharply at the beginning of training and quickly approaches negligible values, providing further evidence of the effectiveness of the learning process. Together, these results confirm that the algorithm is able to approximate the action-value function with increasing accuracy, ensuring that the derived policies converge toward optimal reliable strategies.

Figure~\ref{fig:tt} shows the evolution of the average link travel time across episodes during the learning process. The results reveal a gradual downward trend. It is worth noting that the curve exhibits non-negligible fluctuations around the decreasing trend, which can be attributed to the inherent exploration–exploitation trade-off of reinforcement learning. In particular, the algorithm may select actions that increase the expected travel time if such actions improve the overall reliability of the policy, i.e., the probability of reaching the destination within the given time budget. 

\begin{figure*}
    \centering
    \begin{subfigure}[b]{1\textwidth}
        \centering
        \includegraphics[height=7cm]{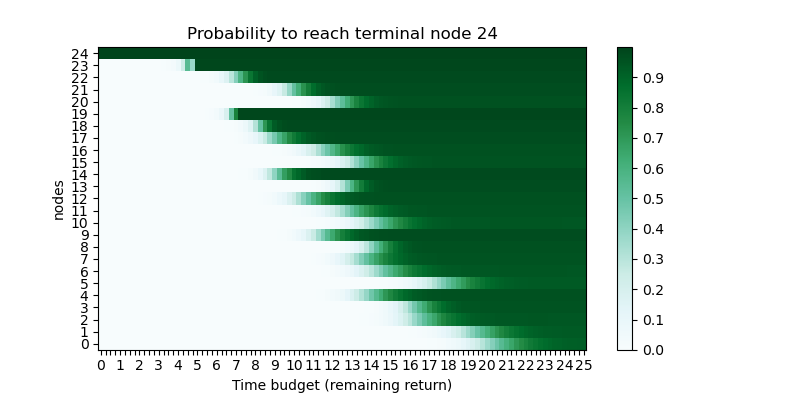}
        \caption{Maximum probability to reach terminal node 24.}
        \label{fig:qf}
    \end{subfigure} \\
    ~ 
    \begin{subfigure}[b]{1\textwidth}
        \centering
        \includegraphics[height=7cm]{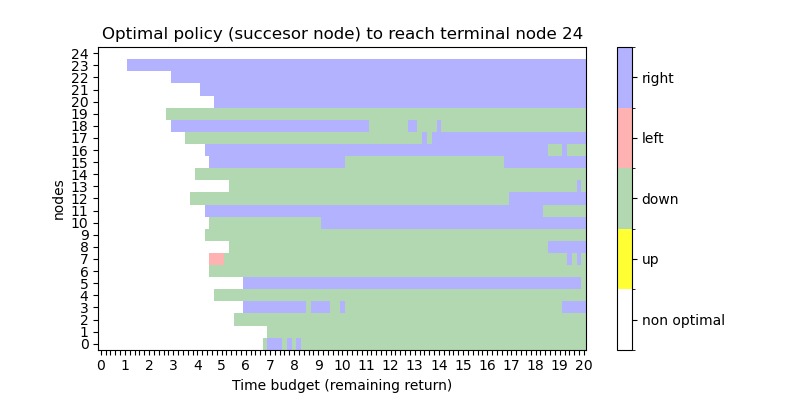}
        \caption{Optimal successor node (policy) to reach terminal node 24.}
        \label{fig:policy2}
    \end{subfigure} \\
    ~ 
    \begin{subfigure}[b]{1\textwidth}
        \centering
        \includegraphics[height=7cm]{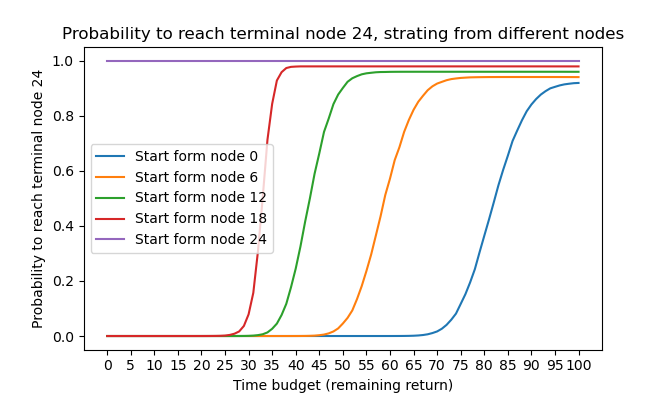}
        \caption{Maximum probability to reach terminal node, starting from different nodes.}
        \label{fig:proba1}
    \end{subfigure} \\
    \caption{Optimal probabilities and optimal policies for the routing problem.}
\end{figure*}

Figure~\ref{fig:steps} presents the evolution of the average number of learning steps per episode throughout training. The results show a slight increase during the initial phase, followed by stabilization around a nearly constant value. This indicates that, as the algorithm progresses, the exploration–exploitation trade-off converges to a stable regime, where learning proceeds with a consistent number of updates per episode. It is important to emphasize that these steps refer to the learning process itself rather than the physical length of the routing paths. When compared with the results in Figure~\ref{fig:tt}, which display a gradual reduction and stabilization of the average travel time, one can observe that improvements in routing performance are achieved without requiring a continuous increase in the number of learning steps. This suggests that the algorithm efficiently refines the value function estimates and the policy while maintaining a stable interaction effort.

Figure~\ref{fig:qf} illustrates the probability of reaching the destination node 24 from each origin node (y-axis) as a function of the available time budget (x-axis). The color scale indicates the probability values, ranging from 0 (white) to values close to 
1 (dark green). As expected, the probability of success increases with the time budget: for small budgets, the probability of reaching the destination is close to zero for most nodes, whereas larger budgets gradually expand the regions of high probability across the network. Furthermore, the figure highlights the dependence of reliability on the starting position: nodes located closer to the destination require smaller budgets to achieve high success probabilities, while nodes farther away demand substantially larger budgets. This visualization provides an interpretable representation of the value function in the reliable routing formulation, explicitly showing how both spatial position and temporal constraints jointly determine the feasibility of reaching the target with high confidence. Overall, the colormap confirms that the proposed reliable reinforcement learning approach effectively captures the trade-off between available time resources and probabilistic guarantees of successful arrival.

Figure~\ref{fig:policy2} depicts the optimal policy for reaching the terminal node~24, where each color represents the optimal action (move up, down, left or right) to take from a given origin node (y-axis) under a specific time budget (x-axis). This figure thus visualizes the structure of the optimal policy $\pi^*(s,t)$ learned by the reliable reinforcement learning algorithm, where $t$ denotes the remaining time budget.
This representation clearly shows that the optimal decision depends jointly on the spatial position (node index) and the temporal constraint (available time budget). For small values of $t$, the policy prioritizes actions that move the agent directly toward the destination along the most time-efficient paths—typically those minimizing expected travel time—since any detour would significantly reduce the probability of timely arrival. As $t$ increases, however, the policy becomes more flexible: it sometimes selects alternative actions corresponding to longer or safer paths that maximize the overall probability of success, given the larger available budget.
This adaptive transition from aggressive, time-minimizing behavior to more reliability-oriented exploration demonstrates a key property of the reliable reinforcement learning framework: its intrinsic capacity to balance efficiency and robustness under uncertainty. The color map provides an interpretable visualization of this trade-off, showing how the learned policy dynamically adjusts its decisions as a function of both spatial location and temporal resources. In regions marked as “non-optimal,” no feasible action guarantees improvement toward the target, reflecting either terminal conditions (starting from node 24) or states excluded by the optimal policy (states with insufficient time budgets).

For example, we can observe from Figure~\ref{fig:policy2} that starting from nodes~20,~21,~22, and~23 with a sufficiently high time budget, the optimal policy $\pi^*(s,t)$ prescribes moving right toward the terminal node~24.
We also notice that the minimum time budget required to ensure optimal reliability increases as the starting node becomes farther from the destination. This illustrates how the policy dynamically adapts to spatial distance by demanding larger temporal resources to maintain the same level of reliability.
Similarly, when starting from nodes~19,~14,~9, and~4 with sufficiently high time budgets, the optimal action is to move down, reflecting the existence of vertically aligned shortest or most reliable routes toward the target.
This directional consistency across symmetric regions of the network highlights the spatial coherence of the learned policy.
An interesting behavior is observed for internal nodes. For example, for node~12: When the remaining time budget $t$ is relatively small (approximately $t < 17$), the optimal policy recommends moving down to node~17, favoring a shorter but potentially riskier route.
However, when $t$ exceeds approximately~17, the optimal decision shifts to moving right to node~13.
This switch indicates that, once sufficient temporal flexibility is available, the agent can afford to take a longer but more reliable path, thereby increasing the overall probability of successful arrival at the terminal node within the allowed time.
Such examples illustrate the adaptive nature of the optimal policy $\pi^*(s,t)$: it explicitly balances path length, reliability, and time budget, demonstrating how temporal resources shape spatial decision-making in the reliable routing problem. 
Overall, Figure~\ref{fig:policy2} highlights how the optimal policy $\pi^*(s,t)$ captures both the temporal and spatial dependencies inherent to stochastic routing problems, offering a clear and interpretable insight into the decision-making process that underlies reliable goal-reaching behavior.

Figure~\ref{fig:proba1} illustrates the evolution of the maximum probability of reaching the destination node (node~24) as a function of the total time budget, when starting from different initial nodes. As expected, the probability of success increases monotonically with the available time budget for all starting positions. Nodes located farther from the destination (e.g., nodes~0 and~6) require significantly larger time budgets to achieve high success probabilities, while nodes located closer to the destination (e.g., nodes~18 and~12) attain near-certain arrival probabilities within shorter time horizons. The curve corresponding to the destination node itself (node~24) remains constant and equal to one, reflecting the trivial case of already being at the terminal state.
The observed sigmoidal shape of the probability curves highlights the trade-off between travel time and reliability: for small time budgets, the probability of timely arrival is close to zero, but once the time budget exceeds a certain threshold—depending on the starting position—the probability rises sharply and eventually saturates near one. This behavior demonstrates that the learned policy effectively captures the stochastic nature of travel times and adapts its routing decisions to maximize the reliability of arrival within the given time constraint. Overall, these results confirm that the reliable reinforcement learning formulation successfully yields interpretable and consistent probability distributions over time budgets, validating the robustness of the approach.
Other complementary illustrations are given in Appendix~\ref{app2}.

\subsection{Price of reliability for the routing problem}

Let us now provide an interpretation of the Q-functions $q^*(i,j,t)$ for the routing problem, where they represent the probability of not exceeding a given time budget $t$. In this formulation, these functions can be viewed as cumulative distribution functions (CDFs), since they give the cumulative probability that the travel time remains below a specified threshold. Unlike the complementary cumulative distribution functions (CCDFs) discussed for the general case (above), which are non-increasing with respect to the threshold variable, these CDF-like functions are non-decreasing with respect to the remaining time budget~$\rho$.

\begin{figure}[h]
        \centering
        \includegraphics[height=7cm]{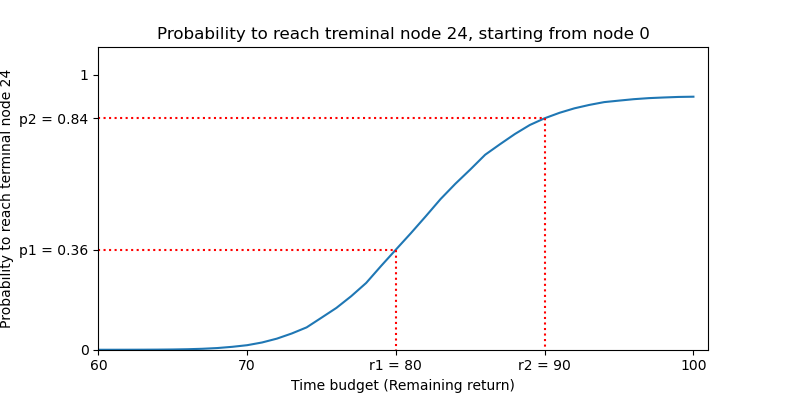}
        \caption{Illustration of the price of reliability principle.}
        \label{fig:por}
\end{figure} 

Figure~\ref{fig:por} illustrates this interpretation by showing the evolution of the maximum probability of reaching the terminal node (node~24) starting from node~0, as a function of the time budget~$t$. As shown, the probability increases monotonically with~$t$, starting from nearly zero for very low time budgets—where reaching the destination is almost impossible—and gradually approaching $1$ as the time budget becomes sufficiently large to guarantee successful arrival. The shape of this curve therefore captures how the reliability of reaching the destination improves as more temporal resources are allocated.

From this figure, one can also derive a quantitative characterization of the trade-off between reliability and time budget. For instance, the probability of reaching the destination within a time budget $t_1 = 80$ is $p_1 = 0.36$, whereas increasing the available time budget to $t_2 = 90$ raises the probability to $p_2 = 0.84$. Hence, an increment of $(t_2 - t_1)$ in time budget corresponds to a gain of $(p_2 - p_1)$ in reliability. This relationship expresses the price of reliability improvement: to increase the probability of success, one must allocate additional time budget, and the required increase depends on the local slope of the CDF.

Conversely, one may interpret this trade-off from the reliability perspective: if the target reliability level is reduced from $p_2$ to $p_1$, the required time budget can be decreased from $t_2$ to $t_1$. Thus, at reliability level $p_2$, the gain $(p_2 - p_1)$ in reliability corresponds to a cost of $(t_2 - t_1)$ in time budget. This bidirectional interpretation emphasizes that the curve encapsulates the fundamental relationship between time allocation and reliability of arrival, which is central to the formulation of reliable and time-constrained routing strategies based on reinforcement learning.

\subsection{Reliable DRL for high-dimensional grid networks}

In this section, we present the results obtained by applying the proposed reliable reinforcement learning (RL) framework to the reliable routing problem in large-scale networks. As discussed previously, the objective is to derive routing policies that maximize the probability of reaching the destination within a predefined time budget, rather than merely minimizing the expected travel time. To address the increased dimensionality and complexity inherent to large networks, we employ a reliable version of the Dueling Double Deep Q-Network (D3QN) algorithm, which combines the stability benefits of Double Q-learning with the representational efficiency of the dueling network architecture. The detailed pseudocode of the adapted reliable D3QN algorithm is provided in Appendix~\ref{RD3QN}, while the corresponding hyperparameters and learning settings used for training are summarized in Table~\ref{tab:dqn}.

The main learning parameters used for the reliable Dueling Double DQN algorithm are summarized in Table~\ref{tab:dqn}. The learning rate was set to $10^{-4}$, providing a balance between convergence stability and learning speed. A discount factor of 
$0.99$ was chosen.
The exploration–exploitation strategy employed a linear decay of the $\varepsilon$-greedy parameter, starting from an initial value of $1$, thereby allowing extensive exploration during the early training stages and a gradual shift toward exploitation as learning progresses. The replay memory buffer was configured with a large capacity of $10^6$ transitions to ensure diverse experience sampling and to reduce temporal correlation between consecutive updates. The mini-batch size was fixed at 
32, which provides a suitable compromise between computational efficiency and the statistical robustness of gradient estimates.

To further enhance stability, the target network was updated every $3\times 10^4$ iterations, combined with a soft update coefficient of $\tau = 10^{-3}$, ensuring smooth and consistent policy updates. The algorithm was implemented with GPU acceleration, enabling efficient parallelization of neural network operations and faster convergence. Overall, these parameter choices follow standard best practices in deep reinforcement learning, ensuring both stability and efficiency while adapting the Dueling Double DQN framework to the requirements of reliable routing in large-scale stochastic networks.

All computations were performed on a workstation equipped with an Apple M3 Max processor featuring 14 CPU cores, a 30-core integrated GPU, and 36~GB of unified memory. This computational setup ensured efficient handling of the extensive simulation data and accelerated the convergence of the deep RL training process. The results presented in the following subsections illustrate the learning behavior, convergence properties, and final performance of the reliable D3QN algorithm when applied to stochastic routing environments of increasing scale and complexity.

\begin{table}[H]
  \centering 
  \caption{Reliable D3QN learning parameters.}
  \label{tab:dqn}
  \begin{tabular}{rl}
     \hline
     Learning rate & 10e-04 \\
     Discount rate & 0.99 \\
     First value of $\varepsilon$ (for $\varepsilon$-greedy policiy) & 1 \\
     Epsilon decay & Linear \\
     Batch size & 32 \\
     Replay memory buffer size  & 10e+06 \\
     Target network upodate frequency & 3 x 10e+04 \\
     Target soft update $\tau$ rate & 1e-03 \\
     DQN algorithm & Reliable D3QN \\
     GPU & Yes \\
     Multi-processing environments & 30 \\
     \hline
  \end{tabular}
\end{table}

\begin{table}[H]
  \centering
  \caption{Computation training time for different grid dimensions. The Maximum time budget as well as the maximum number of steps are
   increased with the increasing of the grid dimension in order to guarantee reliable routes in case of high dimensions.}
  \begin{tabular}{rrrr}
    \hline
    Grid dimension & Max. time & Max. step & Computation time \\
     & budget & number & (training) \\
    \hline
    10 x 10 & 40 & 28.8 M & 43m:42s \\ 
    20 x 20 & 80 & 58.4 M & 1h:17m:24s \\
    40 x 40 & 160 & 250.0 M & 5h:26m:24s \\ 
    50 x 50 & 200 & 517.6 M & 9h:53m:24s \\ 
    60 x 60 & 240 & 800.5 M & 16h:32m:00s \\      
    \hline        
  \end{tabular}
  \label{tab:ct}
\end{table}

Table~\ref{tab:ct} summarizes the computation times obtained when applying the reliable Dueling Double DQN reinforcement learning algorithm to the reliable routing problem on grid networks of increasing dimensions. As the grid size grows from $10 \times 10$ to $60 \times 60$, both the maximum time budget and the maximum number of training steps are proportionally increased in order to ensure the learning of reliable routes over longer spatial distances. Specifically, the maximum time budget is scaled from $40$ for the $10 \times 10$ grid to $240$ for the $60 \times 60$ grid, while the maximum number of training steps rises from $28.8$ million to approximately $800.5$ million. These adjustments are essential to allow the agent sufficient exploration and convergence in larger state spaces, where the number of possible paths grows combinatorially with the grid dimension.

All experiments were conducted in a multi-processing environment using $30$ parallel processes to accelerate experience collection and stabilize learning. As expected, the computation time increases substantially with the grid dimension: from about $43$ minutes for the $10 \times 10$ network to more than $16.5$ hours for the $60 \times 60$ case. This rapid growth highlights the computational cost associated with scaling reliable reinforcement learning algorithms to large, high-dimensional environments. Nevertheless, the observed training durations remain tractable, demonstrating the efficiency of the Dueling Double DQN framework when combined with parallelized training for complex routing tasks.

\section{Conclusion \& Discussion}

Reliable reinforcement learning enables robust and risk-aware decision making in safety-critical domains, from autonomous vehicles, drones, and robotics to healthcare, finance, and energy systems. By explicitly accounting for uncertainty and rare but high-impact risks, it ensures safer, more resilient, and trustworthy control in environments where failures can have severe consequences.

In this work, we introduced a novel formulation of reliable reinforcement learning (RL) that shifts the optimization criterion from maximizing the expected return to maximizing the probability of exceeding a given return threshold. To achieve this, we proposed a state-augmented formulation in which the state space is extended to incorporate the remaining margin to the return threshold. This reformulation allows the reliable RL problem to be cast in the form of a standard RL problem, enabling the direct application of existing RL and deep RL algorithms to derive policies that are optimal in terms of reliability. The theoretical equivalence between the original formulation and its augmented counterpart provides a rigorous foundation for this approach and ensures that reliable strategies can be obtained without the need for entirely new algorithmic designs.

The effectiveness of the proposed formulation was demonstrated in the context of reliable routing problems, where the objective is to maximize the probability of reaching a destination within a given time budget. Numerical experiments using both reliable Q-learning and reliable Dueling Double DQN confirmed that the proposed approach successfully learns reliable policies. The results showed that the algorithms converge, as evidenced by the decreasing error measures, and that the learned policies effectively balance travel efficiency with reliability. Importantly, the experiments highlighted that, unlike standard routing strategies which minimize expected travel time, reliable routing may favor slightly longer paths if they yield a higher probability of reaching the destination within the time budget. This illustrates the essential trade-off between efficiency and reliability that our approach is designed to capture.

Despite these promising results, several limitations deserve discussion. First, the state-augmented formulation enlarges the state space, which may increase computational complexity in large-scale problems, especially when combined with deep RL methods. Second, the current framework assumes full knowledge of transition probabilities or travel time distributions, which may not always be available in real-world applications. Third, while our experiments were conducted on relatively simple network structures, the scalability and robustness of the approach in highly dynamic, large-scale environments remain to be thoroughly investigated.

Future research could address these limitations in several directions. One important line of work concerns the design of efficient function approximation methods to mitigate the curse of dimensionality introduced by the augmented state space. Another promising direction is to integrate model-free or data-driven estimation techniques, enabling the application of reliable RL when the transition dynamics are partially unknown or subject to uncertainty. Extending the framework to multi-agent systems also opens new perspectives, for instance in cooperative routing or distributed control, where agents must coordinate to ensure joint reliability objectives.
Regarding the specific application of the Reliable Reinforcement Learning (R2L) framework to the reliable routing problem, future efforts will focus on incorporating the speed-up and pruning techniques introduced in \citep{Sam12, KSS14} upstream of the R2L procedure. This enhancement aims to reduce computational complexity and improve scalability. The resulting approach will then be benchmarked against existing routing methods that do not rely on reinforcement learning, in order to assess both efficiency and reliability gains.
Finally, combining the proposed approach with risk-sensitive or distributional RL methods may further enhance robustness, offering a principled way to balance expected performance, risk, and reliability in real-world decision-making tasks.

\bibliography{mybib}


\appendix

\section{Complementary illustrations on the application of Q-learning for reliable routing on a 5x5 grid network}
\label{app2}

\begin{figure*}
    \centering
    \begin{subfigure}[b]{1\textwidth}
        \centering
        \includegraphics[height=70mm]{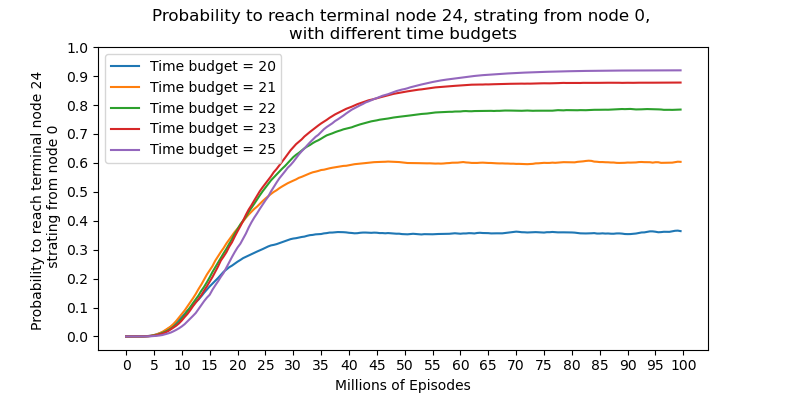}
        \caption{Probability to reach destination node 24, starting from node 0, with different time budgets.}
        \label{fig:probs0}
    \end{subfigure} \\
    ~ \hfill
    \begin{subfigure}[b]{1\textwidth}
        \centering
       \includegraphics[height=70mm]{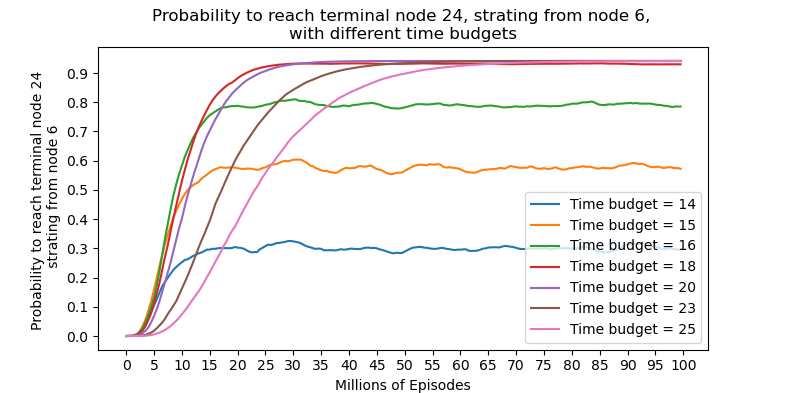}
        \caption{Probability to reach destination node 24, starting from node 6, with different time budgets.}
        \label{fig:probs6}
    \end{subfigure} \\
     ~ \hfill
    \begin{subfigure}[b]{1\textwidth}
        \centering
        \includegraphics[height=70mm]{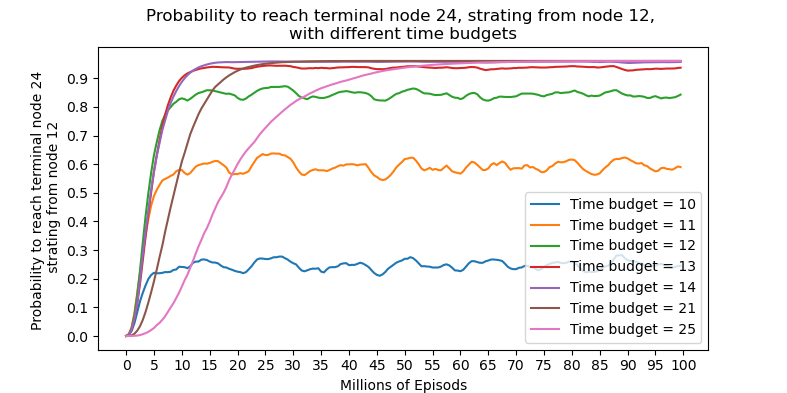}
        \caption{Probability to reach destination node 24, starting from node 12, with different time budgets.}
        \label{fig:probs12}
    \end{subfigure} \\
    \caption{Illustration of the convergence of the reliable Q-learning algorithm for reliable routing.}
\end{figure*}

Figures \ref{fig:probs0}, \ref{fig:probs6}, and \ref{fig:probs12} illustrate the evolution of the maximum probability of reaching the destination node 24 as a function of the number of learning episodes, for several time budgets and for three different starting nodes: 0, 6, and 12 respectively. Across all cases, the curves show a clear upward trend, confirming the progressive improvement of the learned policies as training proceeds. After approximately 40–60 million episodes, the probabilities begin to stabilize, indicating that the learning process converges toward an optimal or near-optimal reliable routing policy.

The dependence on the available time budget is particularly evident. For small budgets (e.g., $T \leq 20$ for node 0 and $T \leq 14$ for node 6), the probability of timely arrival remains below 0.5, as the limited travel time budget is insufficient to reliably reach the destination. As the budget increases, the success probability grows rapidly, reflecting the algorithm’s ability to exploit additional time flexibility to identify more reliable paths. For sufficiently large budgets—typically $T \geq 23$ for node 0 and $T \geq 18$ for node 6—the probability of reaching the destination converges close to 1, demonstrating that the learned policy achieves almost certain arrival within feasible temporal margins.

The comparison among the three figures highlights the spatial consistency of the results. Starting from node 0, the farthest from the destination, convergence requires more learning episodes and yields lower probabilities for the same budgets, due to the longer expected travel distance and cumulative uncertainty along the path. In contrast, when starting from intermediate nodes such as 6 or 12, convergence is faster and final probabilities approach 1 even for moderate time budgets, confirming that the learned Q-function effectively captures both distance and temporal uncertainty in a coherent manner.

These quantitative findings are consistent with the structure of the optimal policy shown in the corresponding policy map. Nodes close to the destination exhibit deterministic and directionally consistent decisions, while nodes located farther away show more varied optimal actions when the time budget is small. This strong alignment between the evolution of reliability probabilities and the spatial distribution of optimal actions provides robust evidence of the validity and interpretability of the proposed reliable reinforcement learning framework for time-constrained stochastic routing.

%
%

\newpage
\section{Reliable Dueling Double DQN algorithm}
\label{RD3QN}

The Reliable Dueling Double DQN (RD3QN) algorithm retains the overall structure of the standard Dueling Double DQN, including the decoupling of action selection and evaluation (double mechanism) and the value–advantage decomposition (dueling head). The key modification lies in the target computation: the state is augmented to incorporate the remaining return needed to exceed the prescribed threshold, and the Bellman update is reformulated accordingly. This adjustment allows the algorithm to optimize the probability of exceeding a return threshold, while leveraging the stability and representational benefits of the original architecture.

\begin{algorithm}
\caption{Reliable Double Deep Q-Network (RD3QN)}
\label{alg:d3qn}
\DontPrintSemicolon
\SetKwInOut{Input}{Input}\SetKwInOut{Output}{Output}
\Input{Discount $\gamma$,  learning rate $\alpha$,  replay capacity $N$, minibatch size $B$, target update period $C$,  exploration schedule $\{\varepsilon_t\}$}
\BlankLine
\textbf{Initialize} replay buffer $\mathcal{D} \leftarrow \emptyset$
\BlankLine
\textbf{Initialize} online network $Q(s, \rho, a;\theta), \quad s\in\mathcal S, \rho \in (\underline{\rho}, \bar{\rho}), a\in \mathcal A, \theta \in \mathbb R$,
with dueling head: \\
$Q(s,\rho, a;\theta) = V(s, \rho;\theta_V) + \big(A(s, \rho, a;\theta_A) - \frac{1}{|\mathcal{A}|}\sum_{a'} A(s,\rho, a';\theta_A)\big)$
\BlankLine
\textbf{Initialize} target network $Q(s,\rho, a;\theta^{-}) \leftarrow Q(s,\rho, a;\theta)$ 
\BlankLine 
\For{episode $=1,2,\dots$}{
  Reset environment and observe $s_0, \rho_0$
  \BlankLine
  \For{$t=0,1,2,\dots$ until terminal}{
    With probability $\varepsilon_t$ select random action $a_t \sim \mathcal{U}(\mathcal{A})$; otherwise
    $a_t \leftarrow \arg\max_{a} Q(s_t, \rho_t,a ;\theta)$ \tcp*[r]{\small $\varepsilon$-greedy}
    Execute $a_t$, observe $r_t$, $s_{t+1}$, and fix $\rho_{t+1} := \max\left(\underline{\rho}, \min(\bar{\rho}, \rho_t - r_t)\right)$
    \BlankLine
    Observe terminal flag $d_t \in \{0,1\}$: $d_t=1$ if $\rho_{t+1}=\underline{\rho}$, otherwise, $d_t=0$ 
    \BlankLine
    Store transition $(s_t,\rho_t,a_t,s_{t+1},\rho_{t+1},d_t)$ in $\mathcal{D}$\; \\
    \If{$|\mathcal{D}| \ge B$}{
      Sample minibatch $\{(s_i,\rho_i,a_i,s'_i,\rho'_i,d_i)\}_{i=1}^B$ from $\mathcal{D}$ (uniform or prioritized)\;
      \tcp*[r]{Double DQN target}
      \ForEach{$i \in \{1,\dots,B\}$}{
        \If{$d_i=1$}{
          $y_i \leftarrow 1$
        }\Else{
          $a^{\star}_i \leftarrow \arg\max_{a} Q(s'_i,\rho'_i,a;\theta)$ \tcp*[r]{\small select with online net}
          $y_i \leftarrow \gamma \, Q(s'_i, \rho'_i, a^{\star}_i; \theta^{-})$ \tcp*[r]{\small evaluate with target net}
        }
      }
      \tcp{Gradient step on squared TD error}
      $\mathcal{L}(\theta) \leftarrow \frac{1}{B}\sum_{i=1}^B \big(y_i - Q(s_i,\rho_i, a_i;\theta)\big)^2$
      \BlankLine
      $\theta \leftarrow \theta - \alpha \nabla_{\theta}\mathcal{L}(\theta)$\;
    }
    \If{$t \bmod C = 0$}{
      $\theta^{-} \leftarrow \theta$ \tcp*[r]{\small hard target update (or use soft update)}
    }
    $s_t \leftarrow s_{t+1}, \; \rho_t \leftarrow \rho_{t+1}$
    \BlankLine
    \If{$d_t=1$}{\textbf{break}}
  }
}
\end{algorithm}

\newpage
\section{Complementary illustrations on the application of DQN for reliable routing on a large grid networks}

\begin{figure}[h]
        \centering
        \includegraphics[width=1\textwidth]{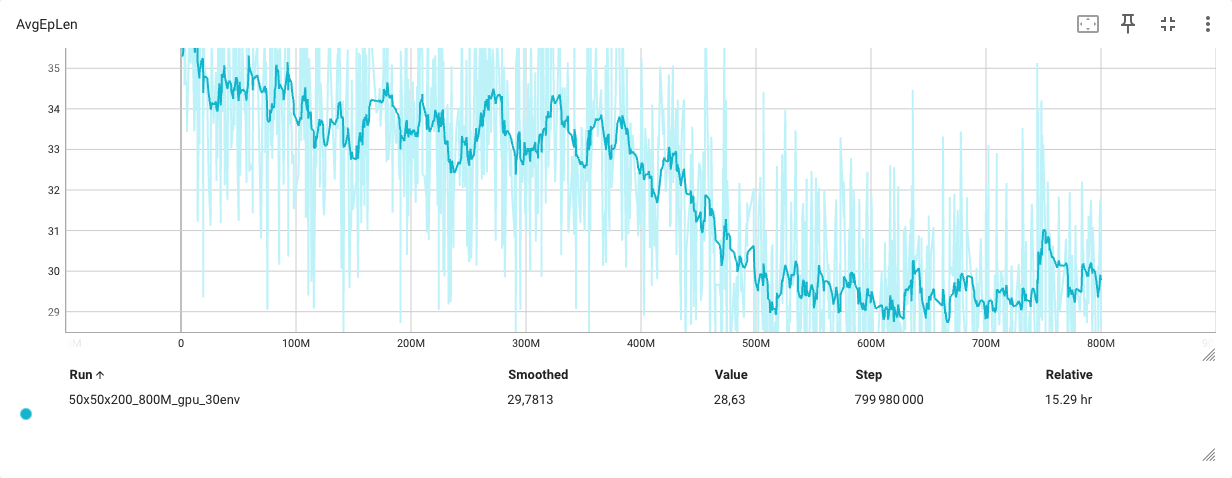}
        \caption{Average episod length. Grid: 50x50, Total time budget: 200.}
        \label{fig:avg_ep_len}
\end{figure}

\begin{figure}[h]
        \centering
        \includegraphics[width=1\textwidth]{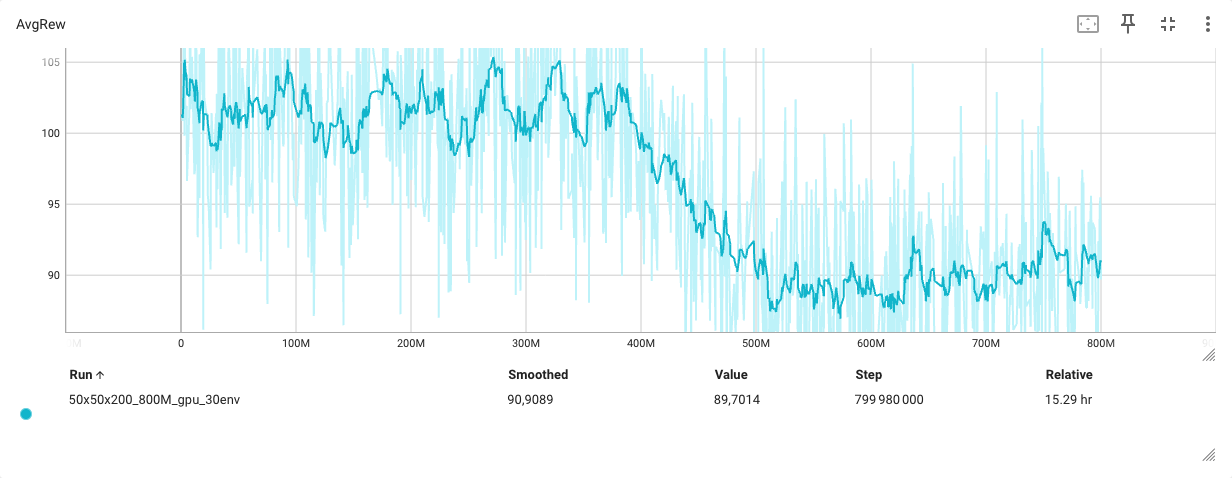}
        \caption{Average link travel time (reward). Grid: 50x50, Total time budget: 200.}
        \label{fig:avg_rew}
\end{figure}

Figures \ref{fig:avg_ep_len} and \ref{fig:avg_rew} illustrate the training performance of the proposed reliable Dueling Double DQN algorithm applied to a 50 x 50 grid network under a total time budget of 200. Figure \ref{fig:avg_ep_len} depicts the evolution of the average episode length over training steps, while Figure \ref{fig:avg_rew} presents the corresponding evolution of the average reward, measured here as the average link travel time. During the initial training phase (up to approximately 400 million steps), both metrics exhibit high variability, reflecting the exploratory behavior of the agent as it samples diverse routing strategies. The episode length and the average link travel time remain relatively high during this period, indicating that the agent is still exploring suboptimal paths and has not yet converged to a stable policy.

As training progresses beyond 400 million steps, both curves demonstrate some stabilization trend. The average episode length decreases, suggesting that the agent increasingly identifies shorter and more efficient routes to the destination. In parallel, the average link travel time also decreases and converges, confirming that the agent learns to minimize the expected travel time while maintaining reliable performance across varying network states. The smoothing of both curves toward the end of training indicates convergence of the Q-values and the stabilization of the learned policy.

Overall, these results confirm the effectiveness of the reliable Dueling Double DQN algorithm in achieving policy convergence for large-scale routing problems. The observed reduction in episode length and average travel time demonstrates the agent’s ability to balance exploration and exploitation efficiently, ultimately learning a stable and near-optimal routing policy adapted to the stochastic nature of the network environment.

\end{document}